\documentclass[11pt]{article}
\usepackage{acl2014}
\usepackage{times}
\usepackage{url}
\usepackage{latexsym}

\usepackage{graphicx}
\usepackage{algorithm}
\usepackage{algorithmic}
\usepackage{subfigure}
\usepackage{multirow}

\usepackage{amsmath}
\usepackage{amsfonts}
\usepackage{amssymb} 
\usepackage{amsthm}

\makeatletter
\def\maketag@@@#1{\hbox{\m@th\normalfont\normalsize#1}}
\makeatother

\usepackage{lipsum}

\title{Resolving Lexical Ambiguity in\\Tensor Regression Models of Meaning}

\author{Dimitri Kartsaklis \\
  University of Oxford \\
  Department of \\Computer Science \\
  Wolfson Bldg, Parks Road \\
  Oxford, OX1 3QD, UK \\
  {\tt {\footnotesize dimitri.kartsaklis@cs.ox.ac.uk}} \\\And
  Nal Kalchbrenner \\
  University of Oxford \\
  Department of \\Computer Science \\
  Wolfson Bldg, Parks Road \\
  Oxford, OX1 3QD, UK \\
  {\tt {\footnotesize nkalch@cs.ox.ac.uk}} \\\And
  Mehrnoosh Sadrzadeh \\
  Queen Mary Univ. of London \\
  School of Electronic Engineering \\
  and Computer Science \\
  Mile End Road \\
  London, E1 4NS, UK \\
  {\tt {\footnotesize mehrnoosh.sadrzadeh@qmul.ac.uk}} 
}

\newcommand{\ov}{\overrightarrow}

\begin{document}
\maketitle

\begin{abstract}
This paper provides a method for improving tensor-based compositional distributional models of meaning by the addition of an explicit disambiguation step prior to composition. In contrast with previous research where this hypothesis has been successfully tested against relatively simple compositional models, in our work we use a robust model trained with linear regression. The results we get in two experiments show the superiority of the prior disambiguation method and suggest that the effectiveness of this approach is model-independent.
\end{abstract}

\section{Introduction}
\label{sec:intro}

The provision of compositionality in distributional models of meaning, where a word is represented as a vector of co-occurrence counts with every other word in the vocabulary, offers a solution to the fact that no text corpus, regardless of its size, is capable of providing reliable co-occurrence statistics for anything but very short text constituents. By \textit{composing} the vectors for the words within a  sentence, we are still able to create a vectorial representation for that sentence that is very useful in a variety of natural language processing tasks, such as paraphrase detection, sentiment analysis or machine translation. Hence, given a sentence $w_1w_2 \dots w_n$, a compositional distributional model provides a function $f$ such that:

\vspace{-0.3cm}
\begin{equation}
  \ov{s} = f(\ov{w_1},\ov{w_2},\dots,\ov{w_n})
  \label{equ:comp}
\end{equation}
\vspace{-0.5cm}

\noindent where $\ov{w_i}$ is the distributional vector of the $i$th word in the sentence and $\ov{s}$ the resulting composite sentential vector. 

An interesting question that has attracted the attention of researchers lately refers to the way in which these models affect ambiguous words; in other words, given a sentence such as ``a man was waiting by the bank'', we are interested to know to what extent a composite vector can appropriately reflect the intended use of word `bank' in that context, and how such a vector would differ, for example, from the vector of the sentence ``a fisherman was waiting by the bank''.

Recent experimental evidence \cite{reddy2011,KartsaklisCONLL,KartsaklisEMNLP} suggests that for a number of compositional models the introduction of a disambiguation step \textit{prior} to the actual compositional process results in better composite representations. In other words, the suggestion is that Eq. \ref{equ:comp} should be replaced by: 

\vspace{-0.65cm}
\begin{equation}
   \ov{s} = f(\phi(\ov{w_1}),\phi(\ov{w_2}),\dots,\phi(\ov{w_n}))
   \label{equ:disamb}
\end{equation}
\vspace{-0.65cm}

\noindent where the purpose of function $\phi$ is to return a disambiguated version of each word vector given the rest of the context (e.g. all the other words in the sentence). The composition operation, whatever that could be, is then applied on these unambiguous representations of the words, instead of the original distributional vectors.

Until now this idea has been verified on relatively simple compositional functions, usually involving some form of element-wise operation between the word vectors, such as addition or multiplication. An exception to this is the work of Kartsaklis and Sadrzadeh \shortcite{KartsaklisEMNLP}, who apply Eq. \ref{equ:disamb} on \textit{partial} tensor-based compositional models. In a tensor-based model, relational words such as verbs and adjectives are represented by multi-linear maps; composition takes place as the application of those maps on vectors representing the arguments (usually nouns). What makes the models of the above work `partial' is that the authors used simplified versions of the linear maps, projected onto spaces of order lower than that required by the theoretical framework. As a result, a certain amount of transformational power was traded off for efficiency.

A potential explanation then for the effectiveness of the proposed prior disambiguation method can be sought on the limitations imposed by the compositional models under test. After all, the idea of having disambiguation emerge as a direct consequence of the compositional process, without the introduction of any explicit step, seems more natural and closer to the way the human mind resolves lexical ambiguities. 

The purpose of this paper is to investigate the hypothesis whether prior disambiguation is important in a pure tensor-based compositional model, where no simplifying assumptions have been made. We create such a model by using linear regression, and we explain how an explicit disambiguation step can be introduced to this model prior to composition. We then proceed by comparing the composite vectors produced by this approach with those produced by the model alone in a number of experiments. The results show a clear superiority of the priorly disambiguated models following Eq. \ref{equ:disamb}, confirming previous research and suggesting that the reasons behind the success of this approach are more fundamental than the form of the compositional function.

\section{Composition in distributional models}
\label{sec:comp}

Compositional distributional models of meaning vary in sophistication, from simple element-wise operations between vectors such as addition and multiplication \cite{Lapata} to deep learning techniques based on neural networks \cite{socher2011,socher2012,nal13cvs}. \textit{Tensor-based models}, formalized by Coecke et al. \shortcite{Coeckeetal}, comprise a third class of models lying somewhere in between these two extremes. Under this setting relational words such as verbs and adjectives are represented by multi-linear maps (tensors of various orders) acting on a number of arguments. An adjective for example is a linear map $f: N \to N$ (where $N$ is our basic vector space for nouns), which takes as input a noun and returns a modified version of it. Since every map of this sort can be represented by a matrix living in the tensor product space $N \otimes N$, we now see that the meaning of a phrase such as `red car' is given by $\overline{red} \times \ov{car}$, where $\overline{red}$ is an adjective matrix and $\times$ indicates matrix multiplication. The same concept applies for functions of higher order, such as a transitive verb (a function of two arguments, so a tensor of order 3). For these cases, matrix multiplication generalizes to the more generic notion of \textit{tensor contraction}. The meaning of a sentence such as `kids play games' is computed as:

\vspace{-0.2cm}
\begin{equation}
   \ov{kids}^\mathsf{T} \times \overline{play} \times \ov{games}
   \label{equ:tensor}
\end{equation}

\noindent where $\overline{play}$ here is an order-3 tensor (a ``cube'') and $\times$ now represents tensor contraction. A concise introduction to compositional distributional models can be found in \cite{KartsaklisSpringer}. 

\section{Disambiguation and composition}
\label{sec:prev}

The idea of separating disambiguation from composition first appears in a work of Reddy et al. \shortcite{reddy2011}, where the authors show that the introduction of an explicit disambiguation step prior to simple element-wise composition is beneficial for noun-noun compounds. Subsequent work by Kartsaklis et al. \shortcite{KartsaklisCONLL} reports very similar findings for verb-object structures, again on additive and multiplicative models. Finally, in \cite{KartsaklisEMNLP} these experiments were extended to include tensor-based models following the categorical framework of Coecke et al. \shortcite{Coeckeetal}, where again all ``unambiguous'' models present superior performance compared to their ``ambiguous'' versions. 

However, in this last work one of the dimensions of the tensors was kept empty (filled in with zeros). This simplified the calculations but also weakened the effectiveness of the multi-linear maps. If, for example, instead of using an order-3 tensor for a transitive verb, one uses some of the  matrix instantiations of Kartsaklis and Sadrzadeh, Eq. \ref{equ:tensor} is reduced to one of the following forms:

\small
\vspace{-0.5cm}
\begin{align}
\begin{split}
\overline{play} \odot (\ov{kids} \otimes \ov{games})~~,~~
  \ov{kids} \odot (\overline{play} \times \ov{games}) \\ 
  (\ov{kids}^{\mathsf{T}} \times \overline{play}) \odot \ov{games}
  ~~~~~~~~~~~~~~~~~~~~~~~
\end{split}  
\end{align}
\vspace{-0.3cm}
\normalsize

\noindent where symbol $\odot$ denotes element-wise multiplication and $\overline{play}$ is a matrix. Here, the model does not fully exploit the space provided by the theoretical framework (i.e. an order-3 tensor), which has two disadvantages: firstly, we lose space that could hold valuable information about the verb in this case and relational words in general; secondly, the generally  non-commutative tensor contraction operation is now partly relying on element-wise multiplication, which is commutative, thus forgets (part of the) order of composition. 

In the next section we will see how to apply linear regression in order to create full tensors for verbs and use them for a compositional model that avoids these pitfalls.

\section{Creating tensors for verbs}
\label{sec:regression}

The essence of any tensor-based compositional model is the way we choose to create our sentence-producing maps, i.e. the verbs. In this paper we adopt a method proposed by Baroni and Zamparelli \shortcite{Baroni} for building adjective matrices, which can be generally applied to any relational word. In order to create a matrix for, say, the intransitive verb `play', we first collect all instances of the verb occurring with some subject in the training corpus, and then we create non-compositional holistic vectors for these elementary sentences following exactly the same methodology as if they were words. We now have a dataset with instances of the form $\langle \ov{subj_i}, \ov{subj_i~play} \rangle$ (e.g. the vector of `kids' paired with the holistic vector of `kids play', and so on), that can be used to train a linear regression model in order to produce an appropriate matrix for verb `play'. The premise of a model like this is that the multiplication of the verb matrix with the vector of a new subject will produce a result that approximates the distributional behaviour of all these elementary two-word exemplars used in training. 

We present examples and experiments based on this method, constructing ambiguous and disambiguated tensors of order 2 (that is, matrices) for verbs taking one argument. In principle, our method is directly applicable to tensors of higher order, following a multi-step process similar to that of Grefenstette et al. \shortcite{GrefSadrBarIWCS13} who create order-3 tensors for transitive verbs using similar means. Instead of using subject-verb constructs as above we concentrate on elementary verb phrases of the form \textit{verb-object} (e.g. `play football', `admit student'), since in general objects comprise stronger contexts for disambiguating the usage of a verb.

\section{Experimental setting}
\label{sec:setting}

Our basic vector space is trained from the ukWaC corpus \cite{ukwac}, originally using as a basis the 2,000 content words with the highest frequency (but excluding a list of stop words as well as the 50 most frequent content words since they exhibit low information content). We created vectors for all content words with at least 100 occurrences in the corpus. As context we considered a 5-word window from either side of the target word, while as our weighting scheme we used local mutual information (i.e. point-wise mutual information multiplied by raw counts). This initial semantic space achieved a score of 0.77 Spearman's $\rho$ (and 0.71 Pearson's $r$) on the well-known benchmark dataset of Rubenstein and Goodenough \shortcite{rubenstein1965}. In order to reduce the time of regression training, our vector space was normalized and projected onto a 300-dimensional space using singular value decomposition (SVD). The performance of the reduced space on the R\&G dataset was again very satisfying, specifically 0.73 Spearman's $\rho$ and 0.72 Pearson's $r$. 

In order to create the vector space of the holistic verb phrase vectors, we first collected all instances where a verb participating in the experiments appeared at least 100 times in a verb-object relationship with some noun in the corpus. As context of a verb phrase we considered any content word that falls into a 5-word window from either side of the verb \textit{or} the object. For the 68 verbs participating in our experiments, this procedure resulted in 22k verb phrases, a vector space that again was projected into 300 dimensions using SVD.

\paragraph{Linear regression} For each verb we use simple linear regression with gradient descent directly applied on matrices $\mathbf{X}$ and $\mathbf{Y}$, where the rows of $\mathbf{X}$ correspond to vectors of the nouns that appear as objects for the given verb and the rows of $\mathbf{Y}$ to the holistic vectors of the corresponding verb phrases. Our objective function then becomes:

\small
\vspace{-0.4cm}
\begin{equation}
 \hat{\mathbf{W}} =  \underset{\mathbf{W}}{\arg\min} \frac{1}{2m} \left( \Vert \mathbf{W} \mathbf{X}^\mathsf{T} - \mathbf{Y}^\mathsf{T} \Vert^2 + \lambda \Vert \mathbf{W} \Vert^2 \right)
\end{equation}
\vspace{-0.4cm}
\normalsize

\noindent where $m$ is the number of training examples and $\lambda$ a regularization parameter. The  matrix $\mathbf{W}$ is used as the tensor for the specific verb.

\section{Supervised disambiguation}
\label{sec:exp1}

In our first experiment we test the effectiveness of a prior disambiguation step for a tensor-based model in a ``sandbox'' using supervised learning. The goal is to create composite vectors for a number of elementary verb phrases of the form \textit{verb-object} with and without an explicit disambiguation step, and evaluate which model approximates better the holistic vectors of these verb phrases. 

\begin{table}[b]
  \small
  \centering
  \begin{tabular}{lll}
   \hline
    \textbf{Verb}  & \textbf{Meaning 1} & \textbf{Meaning 2} \\
   \hline\hline
    break & violate (56) & break (22) \\
    catch & capture (28) & be on time (21) \\
    play  & musical instrument (47) & sports (29) \\
    admit & permit to enter (12) & acknowledge (25) \\
    draw  & attract (64) & sketch (39)  \\
    \hline
  \end{tabular}
  \caption{Ambiguous verbs for the supervised task. The numbers in parentheses refer to the collected training examples for each case.}
  \label{tbl:ambverbs}
  \normalsize
\end{table}

The verb phrases of our dataset are based on the 5 ambiguous verbs of Table \ref{tbl:ambverbs}. Each verb has been combined with two different sets of nouns that appear in a verb-object relationship with that verb in the corpus (a total of 343 verb phrases). The nouns of each set have been manually selected in order to explicitly represent a different meaning of the verb. As an example, in the verb `play' we impose the two distinct meanings of using a musical instrument and participating in a sport; so the first set of objects  contains nouns such as `oboe', `piano', `guitar', and so on, while in the second set we see nouns such as `football', 'baseball'' etc. 

In more detail, the creation of the dataset was done in the following way: First, all verb entries with more than one definition in the Oxford Junior Dictionary \cite{OxfordJun} were collected into a list. Next, a linguist (native speaker of English) annotated the semantic difference between the definitions of each verb in a scale from 1 (similar) to 5 (distinct). Only verbs with definitions exhibiting completely distinct meanings (marked with 5) were kept for the next step. For each one of these verbs, a list was constructed with all the nouns that appear at least 50 times under a verb-object relationship in the corpus with the specific verb. Then, each object in the list was manually annotated as \textit{exclusively} belonging to one of the two senses; so, an object could be selected only if it was related to a single sense, but not both. For example, `attention' was a valid object for the \textit{attract} sense of verb `draw', since it is unrelated to the \textit{sketch} sense of that verb. On the other hand, `car' is not an appropriate object for either sense of `draw', since it could actually appear under both of them in different contexts. The verbs of Table \ref{tbl:ambverbs} were the ones with the highest numbers of exemplars per sense, creating a dataset of significant size for the intended task (each holistic vector is compared with 343 composite vectors).

We proceed as follows: We apply linear regression in order to train verb matrices using jointly the object sets for both meanings of each verb, as well as separately---so in this latter case we get two matrices for each verb, one for each sense. For each verb phrase, we create a composite vector by matrix-multiplying the verb matrix with the vector of the specific object. Then we use 4-fold cross validation to evaluate which version of composite vectors (the one created by the ambiguous tensors or the one created by the unambiguous ones) approximates better the holistic vectors of the verb phrases in our test set. This is done by comparing each holistic vector with all the composite ones, and then evaluating the rank of the correct composite vector within the list of results. 

In order to get a proper mixing of objects from both senses of a verb in training and testing sets, we set the cross-validation process as follows: We first split both sets of objects in 4 parts. For each fold then, our training set is comprised by $\frac{3}{4}$ of set \#1 plus $\frac{3}{4}$ of set \#2, while the test set consists of the remaining $\frac{1}{4}$ of set \#1 plus $\frac{1}{4}$ of set \#2. The data points of the training set are presented in the learning algorithm in random order.

We measure approximation in three different metrics. The first one, accuracy, is the strictest, and evaluates in how many cases the composite vector of a verb phrase is the closest one (the first one in the result list) to the corresponding holistic vector. A more relaxed and perhaps more representative method is to calculate the mean reciprocal rank (MRR), which is given by:

\vspace{-0.3cm}
\begin{equation}
 \text{MRR} = \frac{1}{m} \sum\limits_{i=1}^m\frac{1}{\textit{rank}_i}
\end{equation}
\vspace{-0.3cm}

\noindent where $m$ is the number of objects and \textit{rank}$_i$ refers to the rank of the correct composite vector for the $i$th object.

Finally, a third way to evaluate the efficiency of each model is to simply calculate the average cosine similarity between every holistic vector and its corresponding composite vector.
%
%
The results are presented in Table \ref{tbl:exp1results}, reflecting a clear superiority ($p < 0.001$ for average cosine similarity) of the prior disambiguation method for every verb and every metric.

\begin{table}[t]
  \small
  \centering
  \begin{tabular}{l|cc|cc|cc}
    \hline
     & \multicolumn{2}{|c|}{\textbf{Accuracy}} & \multicolumn{2}{|c|}{\textbf{MRR}} & \multicolumn{2}{|c}{\textbf{Avg Sim}} \\
    \hline\hline
     & Amb. & Dis. & Amb. & Dis. & Amb. & Dis. \\
    \hline\hline
    break & 0.19 & 0.28 & 0.41 & 0.50 & 0.41 & 0.43 \\
    catch & 0.35 & 0.37 & 0.58 & 0.61 & 0.51 & 0.57 \\
    play  & 0.20 & 0.28 & 0.41 & 0.49 & 0.60 & 0.68 \\
    admit & 0.33 & 0.43 & 0.57 & 0.64 & 0.41 & 0.46 \\
    draw  & 0.24 & 0.29 & 0.45 & 0.51 & 0.40 & 0.44 \\
    \hline
  \end{tabular}
  \caption{Results for the supervised task. `Amb.' refers to models without the explicit disambiguation step, and `Dis.' to models with that step.}
  \label{tbl:exp1results}
  \normalsize
\end{table}

\section{Unsupervised disambiguation}
\label{sec:exp2}

In Section \ref{sec:exp1} we used a controlled procedure to collect genuinely ambiguous verbs and we trained our models from manually annotated data. In this section we briefly outline how the process of creating tensors for distinct senses of a verb can be automated, and we test this idea on a generic verb phrase similarity task. 

First, we use unsupervised learning in order to detect the latent senses of each verb in the corpus, following a procedure first described by Sch\"utze \shortcite{Schutze}. For every occurrence of the verb, we create a vector representing the surrounding context by averaging the vectors of every other word in the same sentence. Then, we apply hierarchical agglomerative clustering (HAC) in order to cluster these context vectors, hoping that different groups of contexts will correspond to the different senses under which the word has been used in the corpus. The clustering algorithm uses Ward's method as inter-cluster measure, and Pearson correlation for measuring the distance of vectors within a cluster. Since HAC returns a dendrogram embedding all possible groupings, we measure the quality of each partitioning by using the variance ratio criterion \cite{calinski1974} and we select the partitioning that achieves the best score (so the number of  senses varies from verb to verb). 

The next step is to classify every noun that has been used as an object with that verb to the most probable verb sense, and then use these sets of nouns as before for training tensors for the various verb senses. Being equipped with a number of sense clusters created as above for every verb, the classification of each object to a relevant sense is based on the cosine distance of the object vector from the centroids of the clusters.\footnote{In general, our approach is quite close to the multi-prototype models of Reisinger and Mooney \shortcite{reisinger2010}.} Every sense with less than 3 training exemplars is merged to the dominant sense of the verb. The union of all object sets is used for training a single unambiguous tensor for the verb. As usual, data points are presented to learning algorithm in random order. No objects in our test set are used for training.

We test this system on a verb phase similarity task introduced in \cite{lapata2010}. The goal is to assess the similarity between pairs of short verb phrases (verb-object constructs) and evaluate the results against human annotations. The dataset consists of 72 verb phrases, paired in three different ways to form groups of various degrees of phrase similarity---a total of 108 verb phrase pairs.

The experiment has the following form: For every pair of verb phrases, we construct composite vectors and then we evaluate their cosine similarity. For the ambiguous regression model, the composition is done by matrix-multiplying the ambiguous verb matrix (learned by the union of all object sets) with the vector of the noun. For the disambiguated version, we first detect the most probable sense of the verb given the noun, again by comparing the vector of the noun with the centroids of the verb clusters; then, we matrix-multiply the corresponding unambiguous tensor created exclusively from objects that have been classified as closer to this specific sense of the verb with the noun. We also test a number of baselines: the `verbs-only' model is a non-compositional baseline where only the two verbs are compared; `additive' and `multiplicative' compose the word vectors of each phrase by applying simple element-wise operations.

\begin{table}[t]
  \centering
  \small
  \begin{tabular}{lc}
    \hline
    \textbf{Model} & \textbf{Spearman's $\rho$} \\
    \hline\hline
    Verbs-only &   0.331    \\
    Additive & 0.379 \\
    Multiplicative & 0.301 \\
    Linear regression (ambiguous) & 0.349 \\
    Linear regression (disamb.) & 0.399 \\
    \hline
    Holistic verb phrase vectors & 0.403 \\
    \hline
    Human agreement & 0.550 \\
    \hline
  \end{tabular}
  \caption{Results for the phrase similarity task. The difference between the ambiguous and the disambiguated version is s.s. with $p < 0.001$.}
  \label{tbl:exp2results}
  \normalsize
\end{table}    

The results are presented in Table \ref{tbl:exp2results}, where again the version with the prior disambiguation step shows performance superior to that of the ambiguous version. There are two interesting observations that can be made on the basis of Table \ref{tbl:exp2results}. First of all, the regression model is based on the assumption that the holistic vectors of the exemplar verb phrases follow an ideal distributional behaviour that the model aims to approximate as close as possible. The results of Table \ref{tbl:exp2results} confirm this: using just the holistic vectors of the corresponding verb phrases (no composition is involved here) returns the best correlation with human annotations (0.403), providing a proof that the holistic vectors of the verb phrases are indeed reliable representations of each verb phrase's meaning. Next, observe that the prior disambiguation model approximates this behaviour very closely (0.399) on unseen data, with a difference \textit{not} statistically significant. This is very important, since a regression model can only perform as well as its training dataset allows it; and in our case this is achieved to a very satisfactory level.

\section{Conclusion and future work}
\label{sec:conclusion}

This paper adds to existing evidence from previous research that the introduction of an explicit disambiguation step before the composition improves the quality of the produced composed representations. The use of a robust regression model rejects the hypothesis that the proposed methodology is helpful only for relatively ``weak'' compositional approaches. As for future work, an interesting direction would be to see how a prior disambiguation step can affect deep learning compositional settings similar to \cite{socher2012} and \cite{nal13emnlp}.

\section*{Acknowledgements}

We would like to thank the three anonymous reviewers for their fruitful comments. Support by EPSRC grant EP/F042728/1 is gratefully acknowledged by D. Kartsaklis and M. Sadrzadeh.

\bibliographystyle{acl}
\bibliography{refs}

\end{document}